\documentclass{article}
\usepackage[12pt]{extsizes}
\usepackage[utf8]{inputenc}
\usepackage{times}
\usepackage{amsmath}
\usepackage{amssymb}
\usepackage{amsthm}
\usepackage{natbib}
\usepackage{graphicx}
\usepackage{lipsum}
\usepackage{hyperref}

\usepackage{authblk}
\usepackage{geometry}
\usepackage{adjustbox}
\usepackage{mathtools}
\usepackage{relsize}
\usepackage{multirow}
\usepackage{booktabs}
\usepackage{tabularx}
\usepackage{listings}

\usepackage{xspace}
\usepackage[ruled,vlined]{algorithm2e}

\usepackage{multirow}
\usepackage{xcolor}

\newcommand{\eg}{\emph{e.g.}\xspace}

\newtheorem{definition}{Definition}
\newtheorem{problem}{Problem}
\newtheorem*{problem*}{Problem}

\newcommand{\red}[1]{\color{black} {#1}}

\newcommand{\xhdr}[1]{\vspace{2mm}\noindent{{\bf #1.}}}

\newcommand{\mname}{\texttt{SumGNN}\xspace}
\newcommand{\std}[1]{\scriptsize{$\pm$#1}}

\usepackage{microtype}
\usepackage{subfigure}

\usepackage{etoolbox}
\apptocmd{\thebibliography}{\raggedright}{}{}

\title{\mname: Multi-typed Drug Interaction Prediction via Efficient Knowledge Graph Summarization}

\author[1]{Yue Yu\thanks{Equal Contribution}}
\author[2]{Kexin Huang*}
\author[1]{Chao Zhang}
\author[3]{Lucas M. Glass}
\author[4]{Jimeng Sun}
\author[3]{Cao Xiao}
\affil[1]{College of Computing, Georgia Institute of Technology, Atlanta, GA}
\affil[2]{Health Data Science, Harvard T.H. Chan School of Public Health, Boston, MA}
\affil[3]{Analytic Center of Excellence, IQVIA, Cambridge, MA}
\affil[4]{Department of Computer Science, University of Illinois at Urbana-Champaign, Urbana, IL}

\begin{document}
\maketitle


\abstract{
\noindent\textbf{Motivation:} Thanks to the increasing availability of drug-drug interactions  (DDI)  datasets and large biomedical knowledge graphs (KGs), accurate detection of adverse DDI using machine learning models becomes possible. However, it remains largely an open problem how to effectively utilize large and noisy biomedical KG for DDI detection. Due to its sheer size and amount of noise in KGs, it is often less beneficial to directly integrate KGs with other smaller but higher quality data (e.g., experimental data). Most of existing approaches ignore KGs altogether. Some tries to directly integrate KGs with other data via graph neural networks with limited success. Furthermore most previous works focus on binary DDI prediction whereas the multi-typed DDI pharmacological effect prediction is more meaningful but harder task.

\noindent\textbf{Results:} To fill the gaps, we propose a new method \mname:~{\it knowledge summarization graph neural network}, which is  enabled by a subgraph extraction module that can efficiently anchor on relevant subgraphs from a KG, a self-attention based subgraph summarization scheme to generate reasoning path within the subgraph, and a multi-channel knowledge and data integration module that utilizes massive external biomedical knowledge for significantly improved multi-typed DDI predictions. \mname outperforms the best baseline by up to 5.54\%, and performance gain is particularly significant in low data relation types. In addition, \mname provides interpretable prediction via the generated reasoning paths for each prediction.

\noindent \textbf{Availability:} The code and data are available at \texttt{https://github.com/yueyu1030/SumGNN}.

\noindent\textbf{Contact:} cao.xiao@iqvia.com

\noindent\textbf{Supplementary information:} Supplementary data are available at Bioinformatics online
}

\section{Introduction}
Adverse drug-drug interactions (DDI) are  modifications of the effect of a drug when administered with another drug, which is a common and dangerous scenario for patients with complicated conditions. Undetected adverse DDIs have become serious health threats and caused nearly 74, 000 emergency room visits and 195, 000 hospitalizations each year in the United States alone~\citep{altman13}. To mitigate these risks and costs, accurate prediction of DDIs becomes a clinically important task. Two types of data are being utilized for developing DDI detection models: Manually curated DDI networks and large biomedical knowledge graphs.  

\noindent\textbf{Curated DDI networks:} Researchers have curated DDI networks based on experimental datasets and  literature such as TWOSIDES~\citep{tatonetti2012data}, MINER~\citep{biosnapnets}  and DrugBank~\citep{drugbank,ryu2018deep}.  These curated data are {\it of higher quality but expensive to create and usually smaller in size.} 

\noindent\textbf{Knowledge Graph:} Over the years, large knowledge graph (KG) such as \citep{rotmensch2017learning}, Hetionet~\citep{hetionnet} and DRKG~\citep{drkg2020} have been constructed from literature mining and database integration. However, these  KGs are {\it large and noisy}: out of their tens of thousands of nodes with millions of edges, only a small subgraph is relevant to a prediction target. 

\noindent\textbf{Deep Learning}:  Graph neural networks (GNN)  have achieved great performance by casting DDI prediction as a link prediction problem on DDI graphs~\citep{gysi2020network,zitnik2018modeling,huang2020skipgnn}. However, existing { deep learning} models are often trained  only based on the DDI dataset at hand, ignoring the large biomedical knowledge graph~\citep{drkg2020,hetionnet} which can benefit the DDI predictions since DDI is driven by complicated biomedical mechanism. 
Some recent works~\citep{karim2019drug,KGNN} tried to integrate knowledge graph into the DDI prediction via direct integration of standard KG and GNN methods. But DDI prediction presents unique modeling difficulties since the input KG is large and noisy while the pertinent information for a drug pair is local. Moreover, most existing works also only make binary classification - predicting the presence of DDIs, despite that predicting the particular DDI type is a more meaningful task.

\noindent\textbf{Our Approach}. In this work, we propose a new method \mname that efficiently uses KG to aid drug interaction prediction. \mname enjoys improved predictive performance, efficiency, inductiveness and interpretability. \mname provides the following technical contributions:     

\begin{enumerate}
    \item \textbf{Local subgraph for identifying useful information}. We use local subgraph in the KG around drug pairs to extract useful information, instead of the entire KG. The subgraph formulation allows noise reduction by anchoring on relevant information and is highly scalable since the message passing receptive field is significantly decreased. 
    \item \textbf{Subgraph summarization scheme for generating reasoning path.} We then propose a summarization scheme to generate mechanism pathway for drug interactions. We develop a layer-independent self-attention mechanism to generate signal intensity score for each edge in the subgraph and prune out a KG subgraph pathways that have high scores. As this pruned subgraph is sparse, it provides insights on the biological processes that drive drug interactions. 
    \item \textbf{Multi-channel data and knowledge integration for improved multi-typed DDI predictions}. We propose to use multi-channel neural encoding to aggregate diverse set of data sources, ranging from the summarized subgraph embedding to chemical structures. It enables utilization of massive external biomedical knowledge for significantly improved multi-typed DDI predictions. In addition, the neural encoding takes different subgraph in each propagation, forming an inductive bias that promotes generalizability in low-resource DDI types.
\end{enumerate}

We conduct extensive experiments to show \mname improves DDI prediction significantly. It has up to 5.54\% increase over the best baseline in F1 while the inference time is greatly reduced. Moreover, \mname excels at low-resource settings whereas previous works do not. \mname is also able to provide reasonable clues about the underlying mechanism of the drug interactions.

\section{Related Works}

\xhdr{External knowledge graph integration}
Recently, several efforts have attempted to leverage the KG for downstream tasks such as recommendation~\citep{wang2019kgat,wang2019knowledge,gao2020privacy}, information extraction~\citep{wang2018label,liang2020bond} and drug interaction prediction~\citep{celebi2019evaluation,karim2019drug,KGNN}. \red{For drug interaction prediction,  \cite{takeda2017predicting} integrate the pharmacokinetic (PK) or pharmacodynamic (PD) side-effect when predicting drug interaction and \cite{li2015large} develop a Bayesian network to combine molecular similarity and drug side-effect similarity to predict the drug effect. However, these methods only consider side effect as the external knowledge, which may not be comprehensive enough in our task. 
With the emergence of the biomedical knowledge graphs~\citep{hetionnet,rotmensch2017learning,drkg2020}, more types of entities and relations have been applied to this task, as 
\citet{wang2017predicting,burkhardt2019predicting,celebi2019evaluation,karim2019drug,dai2020wasserstein} project each entity and relation to a dense vector with knowledge graph embedding techniques~\citep{bordes2013translating,su2020network,trouillon2016complex} and then feed them to neural networks for prediction. However, they do not directly harness the neighborhood information for target entities during inference, thus the external knowledge information are not sufficiently exploited. To tackle the above drawback, \citep{KGNN} adopts graph convolutional networks with neighborhood sampling to explicitly model the neighborhood relations with higher inference speed. 
However, as each neighboring entity could play a crucial role in the drug interaction mechanism, random sampling could potentially dropout these important factors and hinders the prediction performance. In contrast, \mname provides a learnable way to extract useful information in the neighborhood. In addition, the previous works all focus on binary DDI prediction whereas \mname evaluates on multi-type relation network. }

\xhdr{Subgraph Graph Neural Network}
Graph neural networks have been proposed for modeling the relation between nodes~\citep{kipf2016semi,velickovic2018graph,schlichtkrull2018modeling,yu2020generalized,srinivasa2020fast}  and have been successfully applied to various domains~\citep{shi2020predicting,yu2020semantic,shang2019pre,yu2020steam}.  
Subgraph structure contains rich information for many graph learning tasks~\citep{teru2019inductive,velickovic2019deep,huang2020graph}. For instance, Ego-CNN applies local ego network to identify structures for graph classification~\citep{egocnn}. \citet{alsentzer2020subgraph} formulate a multi-channel way to subgraph classification. Cluster-GCN~\citep{clustergcn} and GraphSAINT~\citep{zeng2019graphsaint} use subgraphs to improve GNN scalability. More relevant to us,  \citet{zhang2018link}
apply local subgraph for link prediction and GraIL~\citep{teru2019inductive} extend this idea into KG completion task via utilizing multi-relational information. In contrast, \mname is driven by the domain DDI prediction problem and is the first to design a graph summarization module on subgraphs to obtain tractable pathway. It also integrates a new multi-channel neural encoding mechanism. We show these modules significantly improve predictive performance over these related works in Section~\ref{sec:results}.

\section{Method}
\label{method}

\begin{figure*}
    \centering
    \includegraphics[width = 0.9\textwidth]{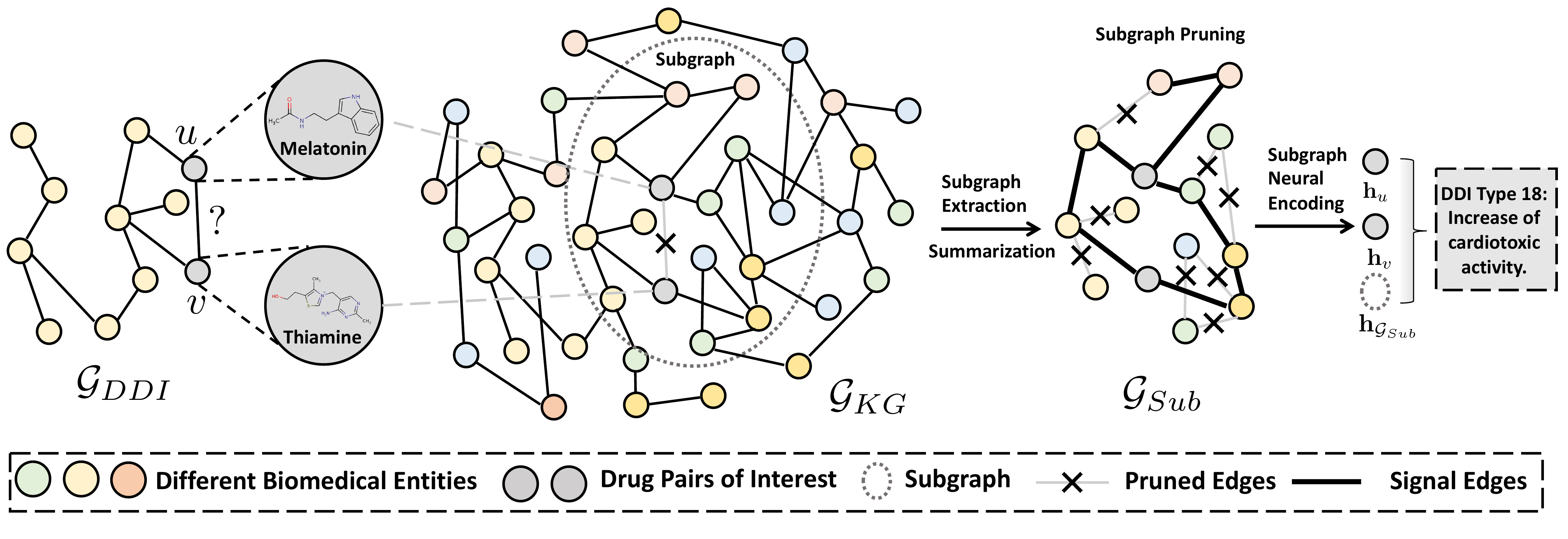}
    \caption{\mname illustration.}
    \vspace{-3mm}
    \label{fig:method}
\end{figure*}


We present \mname in this section\footnote{The \mname code is available at \url{https://github.com/yueyu1030/SumGNN}.}. We summarize problem settings in Section~\ref{definition} and describe our method in detail in Section~\ref{sec:method}. Our method can be decomposed into three modules. First, we extract the local subgraph in the KG around drug pairs to obtain useful information. Then, we propose a summarization scheme to generate a mechanism pathway for drug interactions. After that, we describe a multi-channel neural encoding layer to predict the pharmacological effect.

\subsection{Problem Settings}\label{definition}

\begin{definition}[\textbf{Drug Interaction Graph}]
Given  drugs $\mathcal{D}$ and pharmacological effects $\mathcal{R}_D$,  the drug interaction graph $\mathcal{G}_{DDI}$ is defined as a set of triplets $\mathcal{G}_{DDI} = \{(u, r, v) \mid u \in \mathcal{D}, r \in \mathcal{R}_D, v \in \mathcal{D})\}$, where each triplet $(u, r, v)$ represents that drug $u$ and drug $v$ have pharmacological effect $r$.
\end{definition}

\begin{definition}[\textbf{External Biomedical Knowledge Graph}]
Given a set of various biomedical entities $\mathcal{E}$ and the biomedical relation among the entities $\mathcal{R}$, the external biomedical knowledge graph $\mathcal{G}_{KG}$ is defined as $\mathcal{G}_{KG} = \{(h, r, t) \mid h, t \in \mathcal{E}, r \in \mathcal{R}\}$ with each item $(h, r, t)$ describes a biomedical relation $r$ between entity $h$ and entity $t$. Note that we aggregate the drug entities in $\mathcal{G}_{DDI}$ to $\mathcal{G}_{KG}$, \textit{i.e.}, $\mathcal{R}_{D} \in \mathcal{R}$, and $\mathcal{D} \in \mathcal{E}$.
\end{definition}

\begin{problem}[\textbf{Multi-relational DDI Prediction}]
The Drug-drug interaction (DDI) prediction is to output the pharmacological effect given the a pair of drugs. Mathematically, it is to learn a mapping $\mathcal{F}: \mathcal{D} \times \mathcal{D} \rightarrow \mathcal{R}_D$ from a drug pair $(u, v) \in (\mathcal{D} \times \mathcal{D})$ to the pharmacological effect $r \in \mathcal{R}_D$.
\end{problem}

\subsection{The \mname Method}\label{sec:method}
\mname is composed of three modules: subgraph anchoring, knowledge summarization, and multi-channel neural encoding. For a given drug pair, we anchor to a subgraph of potential biomedical entities that are close to the pairs in the KG. Then, we propose a new graph neural network that has a summarization scheme to provide a condense pathway to reason about the drug interaction mechanism. Given this pathway graph, we use multi-channel neural encoding, to integrate diverse sources of available information to generate a sufficient drug pair representation. At last, a decoding classifier is followed to predict the interaction outcome.
We initialize all entity embedding using KG  method TransE~\citep{bordes2013translating}, where a entity is denoted as $\mathbf{h}_{u}^{(0)}$. 

\subsection*{(A) The Local Subgraph Extraction Module}

The biomedical KG describes the complicated mechanism of human biology. Modulation in several nodes (drug-pairs) in the KG can perturb the connected nodes (\textit{e.g.} disease, cellular component, and etc.) which creates a ripple effect that eventually result in various physiological outcomes~\citep{hetionnet}. The effect is diffused as distance between the drug pairs and the biomedical entities increases. Thus, to understand the drug interactions, we focus on local subgraphs in the KG around the drug pairs. Specifically, for drug pairs $u$ and $v$, we first extract the $k$-hop neighboring nodes for both $u$ and $v$, $\mathcal{N}_k(u) = \{s \mid d(s, u) \leq k\}$ and $\ \mathcal{N}_k(v) = \{s\mid d(s, v) \leq k\}$, where $d(\cdot, \cdot)$ stands for the distance between two nodes on $\mathcal{G}_{KG}$. Then, we obtain the enclosing subgraph based on the intersection of these nodes, $\mathcal{G}_{Sub} = \{(u, r, v) \mid u,v \in \mathcal{N}_k(u) \cap \mathcal{N}_k(v), r \in \mathcal{R}$\}.

Motivated by \citet{zhang2018link} which highlights the importance of node relative position to the central node $u,v$ in the subgraph
, we augment the initial node embedding in the subgraph by concatenating a position vector. For each node $i$ in the subgraph $\mathcal{G}_{Sub}$,  we first compute the shortest path length { $d(i, u)$ and $d(i, v)$} between $i$ and the center drug pairs nodes $u, v$. After that, we convert it into a position vector $\mathbf{p}_i = [\text{one-hot}(d(i, u)) \oplus \text{one-hot}(d(i, v))]$. Then, we update the node $i$ representation as $\mathbf{h}_i^{(0)} = [\mathbf{h}_i^{(0)}, \mathbf{p}_i]$.\\

\vspace{-3mm}
\subsection*{(B) The  Knowledge Summarization Module}

To provide biological insights in addition to the predictive outcome, we design a knowledge summarization module to summarize the subgraph information into a graph-based pathway for potential drug interactions. Note that the pathway is not a linear line, but a sparse subgraph since drug interactions are usually due to complicated interplays among many types of biomedical entities. The summarization requires that we need to retain edges that contain most useful signals for the drug interactions while removing paths that are not important. To achieve this, we adopt a \emph{layer-independent}, \emph{relation-aware} self-attention module to assign a weight for every edge in $\mathcal{G}_{Sub}$. These weights are generated based on the input featurization $\mathbf{h}^{(0)}$ and represent the interaction signal intensities for edge pruning.

Specifically, we denote the interaction signal intensity score for the edge connecting any biomedical entity $i$ and $j$ as $\alpha_{i,j}$. Inspired by the relation-aware transformer architecture~\citep{shaw2018self}, we use self-attention mechanism, which takes account into all neighbor nodes in the subgraph to generate the attention weight. This attention mechanism is ideal for us because it generates the signal intensity score after examining all biomedical entities in the subgraph around the drug-pairs. Here, $\alpha_{u,v}$ is calculated as
\begin{equation}
    \alpha_{i,j} = \operatorname{Threshold}\left (\phi \left(\frac{\mathbf{h}^{(0)}_{j} \mathbf{W}^{J}(\mathbf{h}^{(0)}_{i} \mathbf{W}^{I}+\mathbf{r}_{ij})^{T}}{\sqrt{d}_k}\right), \gamma \right),
    \label{eq:customize}
\end{equation}
where the $\mathbf{W}^J$ and $\mathbf{W}^I$ are the self-attention key weights that contain representation for each node in the subgraph, $r_{ij}$ encodes the relationship between the two entities $i$ and $j$, $\sqrt{d}_k$ is the size of feature vector $\mathbf{h}^{(0)}$ for normalization, $\gamma$ is the signal threshold and $\phi(x) = \frac{e^x-e^{-x}}{e^x+e^{-x}}$ is the \texttt{tanh} function for non-linear transformation.
Intuitively, this function { first computes the dot product between $\mathbf{h}_i^{(0)}$ and $\mathbf{W}^I$ to get an attention score between node $i$ and every neighbor node in the subgraph. Then we sum it up with the relation embedding, followed up the same procedure to calculate attention score between node $j$ and every other node through dot product between $\mathbf{h}_j^{(0)}$ and $\mathbf{W}^J$. By taking the dot product with every other nodes for both $i,j$, the final score considers all the subgraph information. } Then, after non-linear transformation, we calculates the signal intensity score
ranging from $-1$ to $1$ for this edge. At last, we apply a threshold function to screen out edges that are below an intensity score threshold $\gamma$ by setting them with weight 0 since they are not important for the interaction prediction and setting them 0 would prune these edges from message passing process in the graph neural network. This step is applied to every edge in the subgraph. 

Note that  existing graph attention approaches~\citep{velickovic2018graph,cai2020transformer,shaw2018self} generate attention weights for every edge in every layer. However, this way can provide potentially contradicting signals across layers for the same edge, precluding the generation of interpretable pathways. In the contrast, \mname adopts a layer-independent attention mechanism, which only depends on the first layer embedding to prune edges. It provides an unequivocal pathway for model explainability. { As many biological networks are constructed through text mining where many edges are potentially false positives, this pruning mechanism also allows noise reduction.}\\

\vspace{-5mm}
\subsection*{(C) The Multi-Channel Integration Module}
To obtain a powerful representation for drug interaction prediction, we integrate a diverse set of information sources.\\

\noindent\textbf{Channel 1: Summarized Knowledge} Using the knowledge summarization approach we desribed above, we identify a summarized subgraph that is important to input drug pairs. We want to generate the latent representation that leverages this subgraph for the input drug pairs. We integrate it using the following message-passing scheme. For each node $v$, we compute a relation-aware message weighted by the signal intensity score $\mathbf{b}_v^l$ at layer $l$ using
the attention score as
\begin{equation}
    \mathbf{b}_{v}^{(l)}=\sum_{u \in \mathcal{N}_{v}} \alpha_{u,v}^{(l)} \left(\mathbf{h}_{u}^{(l-1)} \mathbf{W}_{r}^{(l)}\right) ,
\end{equation}

\noindent where $\mathcal{N}_{v}$ denotes the neighbors of node $v$ in subgraph $\mathcal{G}_{Sub}$, $\mathbf{W}_{r}^{(l)}$ is the weight matrix to transform hidden representation for node $u,v$'s relation $r$ in layer $l$. To avoid overfitting, we use basis decomposition~\citep{schlichtkrull2018modeling} to decompose $\mathbf{W}_{r}^{(l)}$ into the linear combination of a small number of basis matrices $\{\mathbf{V}_b\}_{b\in B}$ as
\begin{equation}
\mathbf{W}_{r}^{(l)}=\sum_{b = 1}^{B} a_{r b}^{(l)} \mathbf{V}_{b}^{(l)}.
\label{eq:basis}
\end{equation}
Then, we propagate the message $\mathbf{b}_v^{(l)}$ to the updated representation $\mathbf{h}_v^{(l)}$ of node $v$ via
\begin{equation}
\mathbf{h}_{v}^{(l)}=\operatorname{ReLU}\left(\mathbf{W}_{\text {self}}^{(l)} \mathbf{h}_{v}^{(l-1)}+\mathbf{b}_{v}^{(l)}\right),
\end{equation}
\noindent where $\mathbf{W}_{\text{self}}$ is the weight matrix to transform the node embedding itself.\\

\noindent \textbf{Channel 2: Subgraph Features} To obtain the embeddings for subgraphs (denoted as $\textbf{h}_{\mathcal{G}_{Sub}}$), we take the average of all node embeddings in $\mathcal{G}_{Sub}$ at layer $l$ projected by a linear layer as
\begin{equation}
\mathbf{h}_{\mathcal{G}_{Sub}}^{(l)}=\text{Mean} \left(\mathbf{W}_{\text{Sub}}\textbf{h}_i^{(l)}\right).
\label{eq:sf}
\end{equation}
\noindent \textbf{Channel 3: Drug Fingerprint} Molecular information such as chemical fingerprints have shown to be powerful predictor of drug interactions~\citep{huang2020caster}. Thus, in addition to the network representation, we obtain the Morgan fingerprint $\mathbf{f_v}$~\citep{rogers2010extended}, which is a predictive descriptor of drugs, for each drug $v$. Note that it is infeasible to use this feature as the input node feature in the KG since KG consists of various types of nodes other than drugs (\eg Side Effect, Disease and genes) and they cannot be represented by Morgan Fingerprints, which lead to inconsistent node features for GNN propagation. \\

\noindent \textbf{Layer-wise Channels Aggregation}
To assemble various representation generated via each layer, we adopt the layer-aggregation mechanism \citep{xu2018representation}. We concatenate node/subgraph embeddings in every layer, \textit{i.e.}, $\mathbf{h}_v = [\mathbf{h}_v^{(1)}, \mathbf{h}_v^{(2)}, \cdots, \mathbf{h}_v^{(L)}]$ and $\mathbf{h}_{\mathcal{G}_{Sub}} = [\mathbf{h}_{\mathcal{G}_{Sub}}^{(1)},\mathbf{h}_{\mathcal{G}_{Sub}}^{(2)},\cdots, \mathbf{h}_{\mathcal{G}_{Sub}}^{(L)}]$ where is $L$ is the layer size. To integrate chemical fingerprints, we update the layer-aggregated embedding by concatenation of chemical representation: $\mathbf{h}_v = [\mathbf{h}_v \oplus \mathbf{f}_v]$.

At last, we combine the various channels together to obtain the input drug-pairs representation $\mathbf{h}_{u,v} = [\mathbf{h}_u, \mathbf{h}_v, \mathbf{h}_{\mathcal{G}_{Sub}}]$. To predict the relation, we obtain a prediction probability vector $\mathbf{p}_{u,v}$ where each value in the vector corresponds to a the likelihood of a relation. $\mathbf{p}_{u,v}$ is computed via feeding the drug pair representation to a decoder parameterized by $\mathbf{W}_\text{pred}$:
\begin{equation}
\mathbf{p}_{u,v}= \mathbf{W}_{\text{pred}}\textbf{h}_{u,v}.
\end{equation}

\subsubsection{Training and Inference}
During training, for multi-class classification task, we adopt the
the cross entropy loss $\ell_{\text{CE}}$ for each edge $(u, r, v)$ as 
\begin{equation}
    \ell_{\text{CE}}(u, r, v) = -\sum_{r=1}^{R} \mathrm{log}(\hat{y}_r) \cdot y_r ,
\label{eq:ce}
\end{equation}
where $\hat{y}_r = \text{softmax}(\mathbf{p}_{u,v}^{r})=\frac{\exp(\mathbf{p}_{u,v}^{r})}{\sum_{i=1}^{R}\exp(\mathbf{p}_{u,v}^{i})}$ and $y_r$ is the binary indicator if class  $r$ is the correct label for $u$ and $v$. 
For multi-label classification task, given the  edge $(u, r, v)$, we adopt
the binary cross entropy loss $\ell_{\text{BCE}}$ as
\begin{equation}
\ell_{\text{BCE}}(u, r, v)=-\log \hat{y}_{r} -\mathbb{E}_{w \sim P_{w}(v)} \log \left(1-y_{r}^{u, w}\right),
\label{eq:bce}
\end{equation}
where $(u, r, w)$ is the sampled negative edge for relation $r$. This is achieved by replacing node $v$ to node $w$ that is sampled randomly according to a distribution $P_{w}(v)\propto d_{w}(v)^{3 / 4}$~\citep{mikolov2013distributed}. Then $\hat{y}_{r}=\text{sigmoid}(\mathbf{p}_{u,v}^{r})$, $y_{r}^{u w}=\text{sigmoid}(\mathbf{p}_{u,w}^{r})$ is the prediction score for two edges. 
Considering all edges, the final loss  $\mathcal{L}$ in \mname is
\begin{equation}
\mathcal{L} = \sum_{(u, r, v) \in \mathcal{E}}{\ell}(u, r, v),
\end{equation}
where $\ell$ is either Eq.~\eqref{eq:ce} or Eq.~\eqref{eq:bce} depending on the task type. During training, we learn the model parameter by minimizing
the total loss $\mathcal{L}$ using stochastic gradient optimizers such as Adam~\citep{kingma2014adam}.

During inference, an unseen node pair $u,v$'s subgraph in the KG is extracted and fed into the same pipeline to calculate the relation vector. For multi-class task, we use the highest probability relation as the predicted relation and for multi-label task, we collect all scores from both positive and negative counterparts for all relations.

\section{Experiments}
\label{experiment}

\subsection{Experiment Setup}
\textbf{Datasets}
(1) \underline{DrugBank} dataset~\citep{drugbank} contains 1,709 drugs (nodes) and 136,351 drug pairs (edges), which are associated with 86 types of pharmacological relations between drugs, such as increase of cardiotoxic activity, decrease of serum concentration and etc. Each drug pair can contain one or two relations. As more than 99.8\% of edges have only one edge type~\citep{ryu2018deep}, we filtered the edge with more than one type in our study. (2) \underline{TWOSIDES}~\citep{tatonetti2012data} dataset contains 645 drugs (nodes) and 46,221 drug-drug pairs (edges) with 200 different drug side effect types as labels. For each edge, it may be associated with multiple labels. Following~\citep{zitnik2018modeling,dai2020wasserstein}, we keep 200 \emph{commonly-occurring} DDI types ranging from Top-600 to Top-800 to ensure every DDI type has at least 900 drug combinations.  (3) For external knowledge base, we use \underline{HetioNet}~\citep{hetionnet}, which is a large heterogeneous knowledge graph merged from 29 public databases. To \textbf{ensure no information leakage}, we remove all the overlapping DDI edges between HetioNet and the dataset. 
In the end, we obtain 33,765 nodes out of 11 types (e.g., gene, disease, pathway, molecular function and etc.) with 1,690,693 edges from 23 relation types. \\

\noindent \textbf{Baselines} We compare our models with several baselines\footnote{Further details on baseline methods, implementation and parameters are in the supplementary.}. 
\begin{itemize}
    \item \underline{MLP}~\citep{rogers2010extended} uses a two-layer MLP on Morgan fingerprint to directly predict drug interactions.
    \item \underline{Deepwalk}~\citep{deepwalk} first learns the embeddings for drugs in the network via random walk. Then, it predict the relation for drug pairs via a linear layer over embeddings. 
     \item \underline{LINE}~\citep{tang2015line} use a one-layer feedforward neural network to learn the embeddings for drugs. Then, it stack a linear layer over embeddings to predict the relation.
    \item \underline{Node2vec}~\citep{grover2016node2vec} first learns the embeddings for drugs in the network. Similar to Deepwalk, it predict the relation via a linear layer over embeddings. 
    \item  \underline{Decagon}~\citep{zitnik2018modeling} adopts multi-relational graph convolutional network~\citep{schlichtkrull2018modeling} on the DDI network for drug interaction prediction.
    \item  \underline{GAT}~\citep{velickovic2018graph} uses attention networks to aggregate neighborhood information in DDI network. 
    \item \underline{SkipGNN}~\citep{huang2020skipgnn} predicts drug interactions by aggregating information from both direct interactions and second-order interactions via two GNNs.
    \item PRD~\citep{wang2017predicting} first use KG embeddings for drugs in the KG, then pass through a linear layer for drug interaction prediction.
    \item \underline{KG-DDI}~\citep{karim2019drug} first extracts KG embeddings for drugs in the KG, then adopts a Conv-LSTM model using the embeddings for drug interaction prediction.
    \item \underline{GraIL}~\citep{teru2019inductive} is for inductive relation prediction on KGs, which uses local subgraph.
    \item \underline{KGNN}~\citep{KGNN} samples and aggregates neighborhoods for each node from their local receptives via GNN and with external knowledge graph, which achieves the state-of-the-art result on binary DDI prediction problem.
\end{itemize}

\noindent\textbf{Metrics}
The task on the DrugBank dataset is a multi-class classification, thus we consider the following metrics:
\begin{itemize}
    \item \underline{F1 Score}: average F1 score over \emph{different classes} as  $\text{F1 Score} = \frac{1}{N} \sum_{k=1}^{N} \frac{2P_k \cdot R_k}{P_k+R_k},$ where $N$ is the \# of classes and $P_k, R_k$ is the precision and recall for $k$-th class. Since it gives equal weights for each classes, they are \emph{more sensitive} to the results for classes where  samples are fewer.
    \item \underline{Accuracy}: 
    Accuracy over all samples  
    $\text{Accuracy} = \frac{|Y_{k} \cap \hat{Y}_{k}|}{|Y_{k}|}$,
     $Y_k$ is the predicted labels at $k$ and $\hat{Y}_k$ are the ground-truth labels.
    \item \underline{Cohen's Kappa}~\citep{cohen1960coefficient} measures the inter-annotator agreement as $\kappa=(p_{o}-p_{e})/(1-p_{e}),$ where $p_{o}$ is the observed agreement (identical to accuracy), $p_{e}$ is the probability of randomly seeing each class.
\end{itemize}
The task on the TWOSIDES dataset is a multi-label prediction. We  follow~\citep{zitnik2018modeling} and consider the following measure. 
For each side effect type, we calculate the performance individually  and use the average performance over all side effects as the final result. 
\begin{itemize}
    \item \underline{ROC-AUC} is the average area under the receiver operating characteristics curve as $\text{ROC-AUC}=\sum_{k=1}^{n} \operatorname{TP}_{k} \Delta \operatorname{FP}_{k},$ where $k$ is $k$-th true-positive and false-positive operating point $(\operatorname{TP}_{k}, \mathrm{FP}_{k})$.
    \item \underline{PR-AUC} is the average area under precision-recall curve $\text{PR-AUC}=\sum_{k=1}^{n} \operatorname{Prec}_k \Delta \operatorname{Rec}_k$
where $k$ is $k$-th precision/recall operating point $(\operatorname{Prec}_k, \operatorname{Rec}_k)$.
    \item \underline{AP@50} is the average precision at 50, where 
    $\text{AP@}k = \frac{|Y_{k} \cap \hat{Y}_{k}|}{|Y_{k}|}$,
     $Y_k$ is the predicted labels at $k$ and $\hat{Y}_k$ are the ground-truth labels.
\end{itemize}

\noindent\textbf{Evaluation Strategy.} For both datasets, we split it into 7:1:2 as train, development and test set. For the \underline{DrugBank} dataset, since the label distribution is highly imbalanced, we ensure train/dev/test set contain samples from all classes.
For the \underline{TWOSIDES} dataset, we use the same method in~\citep{zitnik2018modeling} to generate negative counterparts for each positive edge by sampling the complement set of positive examples. For every experiment, we conduct five independent runs and select the best performing model based on the loss value on the validation set. 

\begin{table}[t]
    \centering
    \caption{\mname achieves the best predictive performance compared to state-of-the-art baselines in DDI prediction. Average and standard deviation of five runs are reported. For these metrics, higher values always indicate better performance. }
    \label{tab:main}
    \begin{adjustbox}{max width=0.95\textwidth}
    \begin{tabular}{l|ccc|ccc}
    \toprule
    Dataset &\multicolumn{3}{c|}{\textbf{Dataset 1: DrugBank}} & \multicolumn{3}{c}{\textbf{Dataset 2: TWOSIDES}} \\ \midrule
    Classification Task &\multicolumn{3}{c|}{Multi-class} & \multicolumn{3}{c}{Multi-label} \\ \midrule
    Methods & F1 Score & Accuracy & Cohen’s Kappa & ROC-AUC & PR-AUC & AP@50\\
    \midrule
    MLP~\citep{rogers2010extended} & 61.10\std0.38 & 82.14\std0.33 &80.50\std0.18& 82.60\std0.26 &	81.23\std0.14&	73.45\std0.28  \\
    Deepwalk~\citep{deepwalk} &\red{24.77\std 0.40}  &  \red{68.50\std0.38} & \red{58.44\std0.23}& \red{88.27\std 0.09}& \red{85.42\std0.14} &\red{81.09\std 0.16} \\
    LINE~\citep{tang2015line}  &\red{30.26\std 0.45}  &  \red{76.57\std0.49} & \red{70.91\std0.53}& \red{91.20\std 0.34}& \red{90.02\std0.28} &\red{84.07\std 0.19} \\
    Node2Vec~\citep{grover2016node2vec} &24.92\std0.32  &  71.09\std0.40 & 63.79\std0.37& 90.66\std 0.13& 88.87\std0.23 &83.00\std 0.30 \\
    Decagon~\citep{zitnik2018modeling} &57.35\std0.26 &87.19\std0.28& 86.07\std0.08 &91.72\std0.04 &	90.60\std0.12	& 82.06\std0.45\\
    GAT~\citep{velickovic2018graph} &33.49\std0.36&77.18\std 0.15 & 74.20\std0.23 & 91.18\std0.14&	89.86\std0.05 &	82.80\std0.17  \\
    SkipGNN~\citep{huang2020skipgnn} &59.66\std0.26& 85.83\std0.18& 84.20\std0.16 &92.04\std0.08 &	90.90\std0.10	&84.25\std0.25 \\
    \midrule
    \red{PRD~\citep{wang2017predicting}} & \red{40.73\std0.44}& \red{81.52\std 0.34}& \red{78.80\std 0.36} & \red{88.75\std0.23}& \red{85.26\std0.33} & \red{79.86\std0.26} \\
    KG-DDI~\citep{karim2019drug} & 36.39\std0.32 &82.48\std0.12& 78.89\std0.27&90.75\std0.07& 88.16\std0.12 & 83.48\std0.05\\
    GraIL~\citep{teru2019inductive} &81.31\std0.30 & 89.89\std 0.24& 88.07\std0.20 &92.89\std 0.09& 91.10\std 0.19 & 86.21\std0.05 \\
    KGNN~\citep{KGNN} &73.99\std0.11& 90.89\std0.20& 89.64\std0.24&92.84\std0.07	&90.78\std0.20&	86.05\std0.12\\ 
    \midrule
    \mname (Ours) &\textbf{86.85\std0.44}&\textbf{92.66\std 0.14}& \textbf{90.72\std0.13} &\textbf{94.86\std 0.21} &	\textbf{93.35\std 0.14} &	\textbf{88.75\std 0.22}\\ \midrule
    \mname-KG & 78.35\std0.51& 89.05\std0.36&87.28\std0.08  & 92.62\std0.13&90.80\std0.40 & 85.75\std0.10\\
    \mname-Sum (w/o Summarization) & 83.20\std0.34& 90.83\std0.19&  90.14\std0.10&94.09\std 0.16& 92.55\std0.24& 87.65\std0.24 \\
    \mname-SF (w/o Subgraph Features) & 84.47\std0.22 & 91.88\std 0.21& 90.26\std0.19 &93.94\std0.11 &92.45\std0.22 &87.69\std0.08 \\
    \mname-CF (w/o Chemical Features) & 83.57\std0.36 & 91.31\std0.17& 90.07\std 0.11 &94.35\std0.11& 92.86\std 0.20& 88.10\std0.07 \\
    \mname-LIA (w/o Layer Independent Attention) & 86.54\std0.22 & 92.44\std0.30& 90.34\std 0.10 &93.92\std0.31& 92.33\std 0.19& 86.15\std0.13 \\
    \bottomrule
    \end{tabular}
    \end{adjustbox} 
\end{table}

\subsection{\mname achieves superior predictive performance} ~\label{sec:results}

We report the performance of our model and all baselines in table~\ref{tab:main}. From the result, we find that \mname achieves the best performance in DDI prediction on two datasets, accurately predicting the correct DDI pharmacological effect consistently. Particularly, \mname has 27.19\%, 5.47\%, 4.65\% absolute increase over the best baseline without KG on three metrics respectively on DrugBank dataset and 2.84\%, 2.45\%, 4.50\% increase on TWOSIDES dataset. Also, \mname achieves 5.54\%, 1.77\%, 1.08\% and 1.97\%, 2.25\%. 2.54\% absolute increase over the state-of-the-art baselines with KG on two datasets. These results clearly verifies the efficacy of our method. 

\subsection{\mname excels at {low-data imbalanced} relations predictions} 

\begin{figure*}[t]
  \centering
  \subfigure[The number of training samples wrt. different classes in DrugBank dataset.]{
    \includegraphics[width=0.45\textwidth]{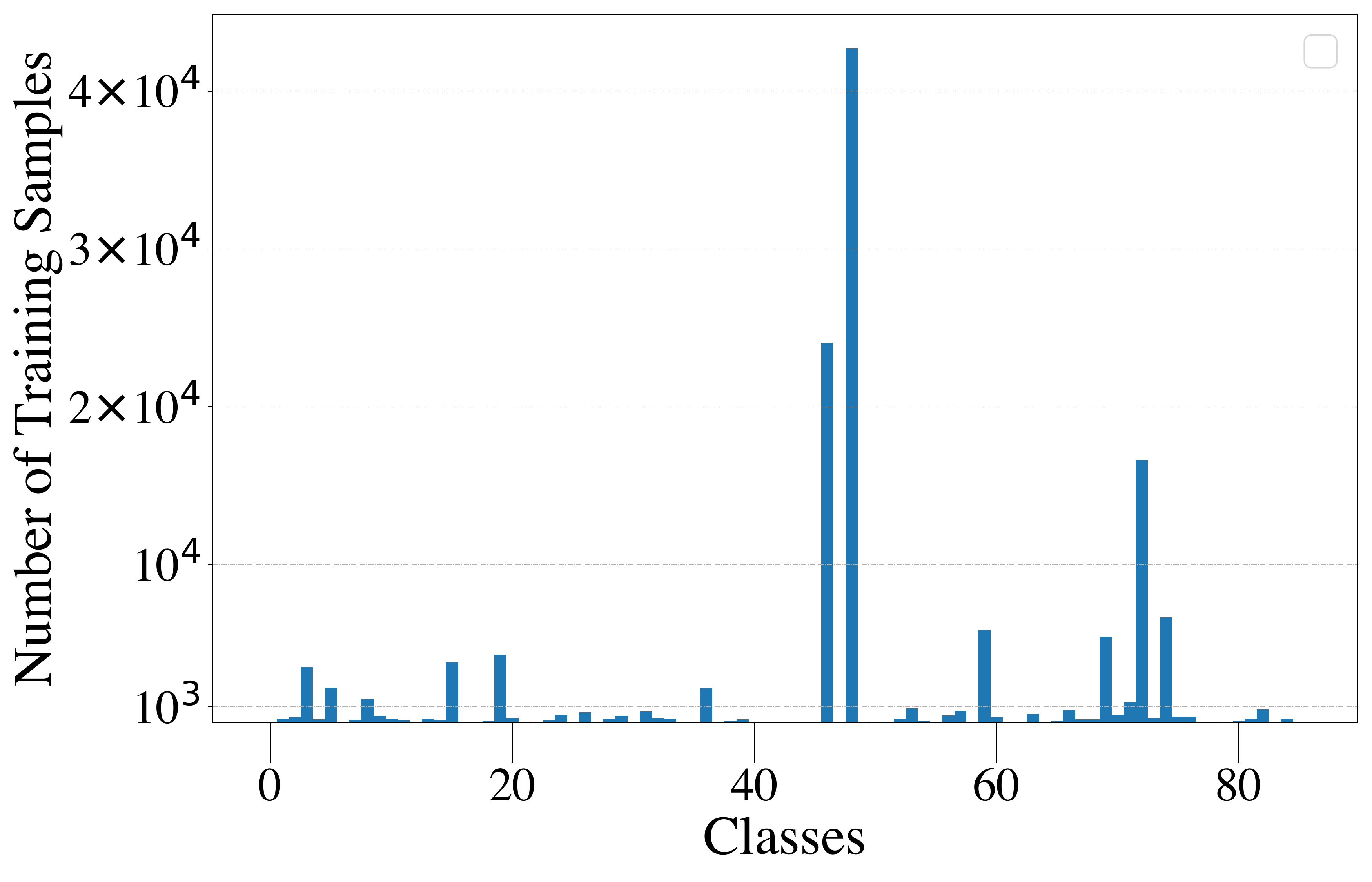}
    \label{fig:dis}
  }\hspace{0.2mm}
  \subfigure[The average F1 score vs. different size of training samples.]{
    \includegraphics[width=0.45\textwidth]{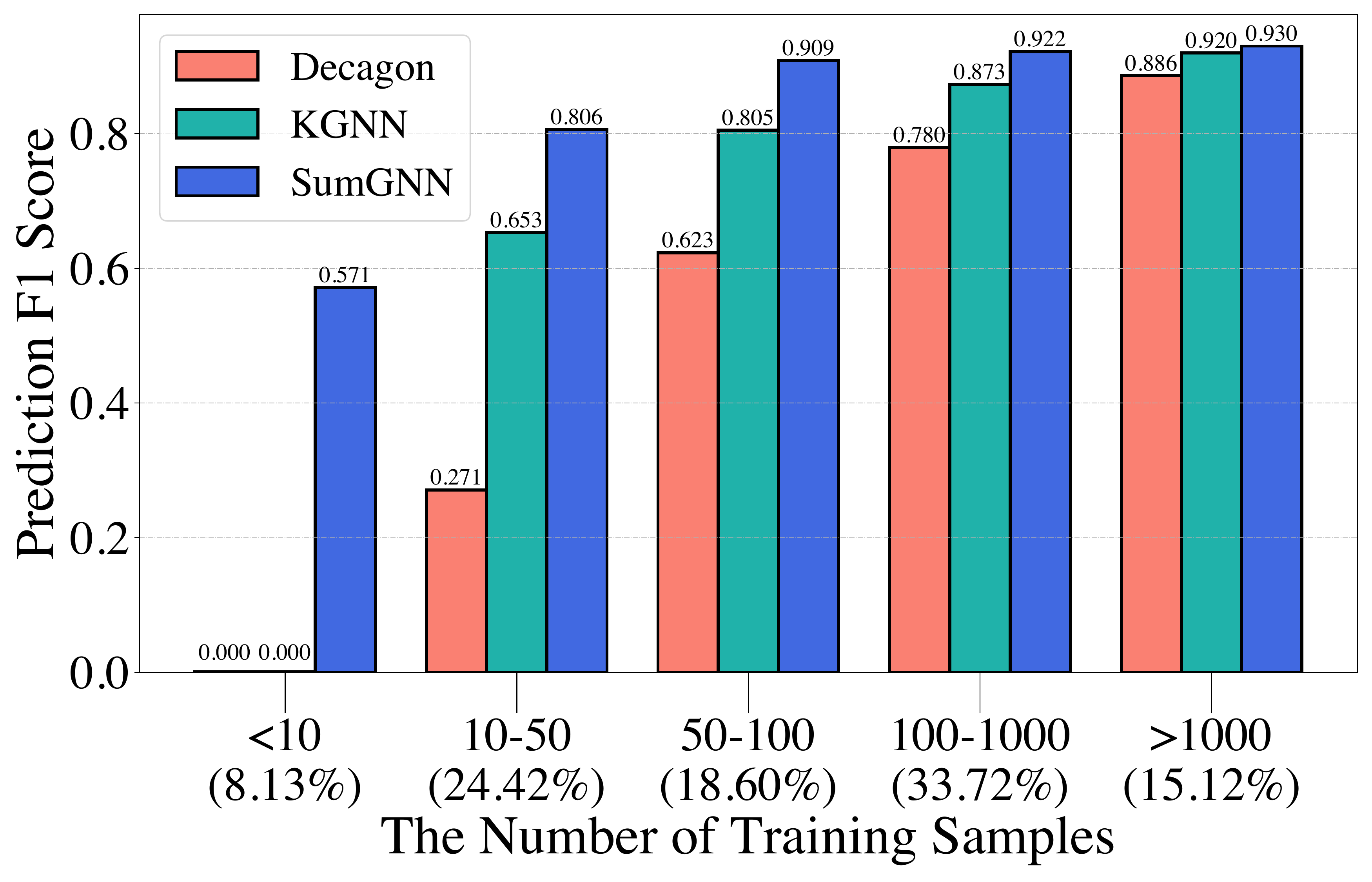}
    \label{fig:flip}
  }
  \vspace{-1mm}
    \caption{The dataset statistics and the average F1 score for different relations with various number of training samples on DrugBank dataset. Here, in (a), the x-axis indicates the index of classes (there are 86 classes in total) and the y-axis indicates the number of training samples in the corresponding class. 
    in (b), the x-axis indicate the range for number of training samples in this group as well as the proportion of classes in this group to all classes and  the y-axis stands for the average prediction F1 score for the classes in the group.}
    \label{fig:label}
\end{figure*}

We observe that the improvement of \mname is much \emph{more significant} on DrugBank than TWOSIDES dataset. We take a closer look at this problem and find that the major difference of these two datasets is that the data distribution of DrugBank is \emph{more imbalanced} -- more than 30\% of the relation types occurs less than 50 times in the training set while more than 10\% of the relation types occurs more than 1000 times, as shown in Fig.~\ref{fig:dis}. 

To examine the model's prediction performance on the size of training data associated with each relation, we first split the relations types into 5 groups with various amounts of training data and then plot the average F1 score of these bins in Figure~\ref{fig:label}. By comparing the performance of \mname against the strongest baseline on Accuracy (KGNN) and the variant of no KG (Decagon), we find that \mname can effectively boost the performance when the samples are extremely scarce. When the size of the training samples of the relation is less than 10, both decagon and KGNN cannot give any correct predictions, while our model can still achieve 57.14\% F1 score (i.e., choose the correct relation out of the 86 relations). { One potential reason is that \mname feeds different subgraphs in every GNN propagation, which forms a much-needed inductive bias over unseen subgraphs, such as the ones in the low-data relations.} This is in sharp contrast to previous approaches such as KGNN. It also justifies that \mname's knowledge summarization via subgraphs is more effective to harness the external knowledge. In addition, we find \mname also brings at least 38.19\% improvement on F1 score for relations occuring less than 50 times. This is an important finding since a high overall accuracy does not mean good performance across all relations. The low-data relations are the hardest to predict correctly and we show that \mname can be used for these low-data relations whereas other previous models cannot. 


\subsection{External knowledge improves prediction}

\red{We find that the first seven method cannot obtain satisfactory result compared to {\mname}. MLP, Deepwalk, LINE and Node2vec only use shallow embedding layers to learn the features for drugs while do not use graph neural networks for modeling the drug interactions, which have limited expression power. Moreover, although GAT considers the attention on different edges, it fails to consider the multi-relation information, as its performance is also not good. 
By comparing the latter four methods with the former five methods, we find the use of KG significantly improves DDI performance, highlighting the necessity of combining external knowledge usage with graph neural networks, as it can provide complementary information for DDI task. }

\subsection{Knowledge summarization module is the best to capture external knowledge} 

Comparing \mname with KGNN and KG-DDI, we show that our model can consistently outperform them on both datasets (more than 4\% on average for DrugBank and 2\% on average for TWOSIDES), which indicates that  simply adopting KG embeddings as well as neighborhood sampling are \emph{insufficient to fully harness the KG information} for DDI prediction. \mname provides the best approach to leverage the external KG and also corroborates with our motivation that the use of subgraph reduces noise and irrelevant information. Although GraIL also extracts subgraphs for downstream tasks, it does not use any knowledge summarization techniques, which potentially further eliminating the irrelevant information in the local subgraph. In addition, GraIL merely considers the position information while neglecting the multi-channel features during information propagation.

\vspace{-3mm}
\subsection{Ablation Study}

To study the usage of KG, we remove the knowledge graph (\mname-KG) and perform prediction on the DDI graph. We see \mname has 8.5\% absolute increase in Macro F1 on DrugBank, highlighting the usefulness of KG. To evaluate the knowledge summarization module, we remove the summarization component (\mname-Sum) and use the raw local subgraph to predict the outcome. We see \mname has 3.65\% increase for DrugBank and 2.24\% increase for TWOSIDES on Macro F1, suggesting that the summarization further condenses the relevant information and elevate the performance.

To study the effect of multi-channel neural encoding, we compare the result of \mname with several variants that remove specific channel information (i.e. subgraph features and chemical features), and we find that these channel information all contribute to the overall performance. Particularly, after removing the subgraph embedding (\mname-SF), the Macro F1 drops by 2.38\% on DrugBank and 0.92\% on TWOSIDES respectively. When removing chemical fingerprint (\mname-CF), the performance also degrades 3.28\% on Macro F1 for DrugBank, corroborating with the indispensable roles of each of the channel. We also see that by replacing the layer independent attention (\mname-LIA) with the previous layer-dependent ones, the performance drops. This suggests that our attention mechanism not only provides interpretability but also increases predictive performance. 

\vspace{-1.5mm}
\subsection{Case Study}

\begin{figure}
    \centering
    \includegraphics[width = 0.5\textwidth]{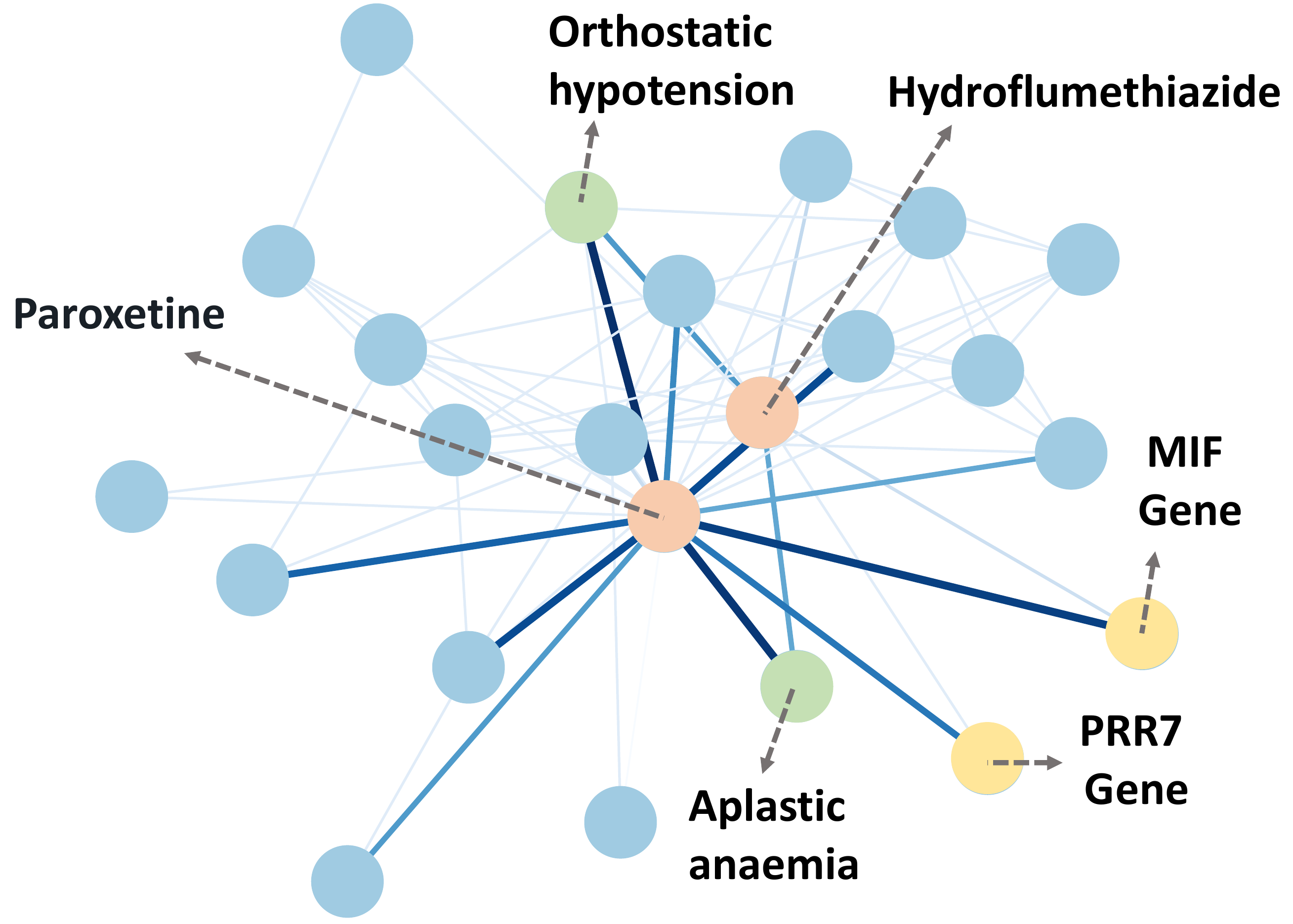}
    \caption{\mname generates a short reasoning path to provide clues for understanding drug interactions. The shade of color indicates the strength of attention weight. Low-weight edges in the extracted subgraph are pruned by \mname and \mname focus on a sparse set of signal edges and nodes. }
    \label{fig:casestudy}
\end{figure}

{ The usefulness of this model lies in twofolds. First, given the high predictive performance, it can identify novel drug-drug interactions that are flagged high by the model while not in the dataset. Second, using the external knowledge summarization module, we are able to discover signal edges, which provide biological pathways to hint at the potential mechanism of DDIs. We provide a case study of a novel drug pair predicted by the model, Paroxetine and Hydroflumethiazide. Paroxetine is an antidepressant, and Hydroflumethiazide is used to treat hypertension and edema. \mname assigns highest probability for the DDI type "increase of the central nervous system depressant activities". We then visualize the generated pathway from \mname's summarization scheme in Fig~\ref{fig:casestudy}. We see that the model significantly reduces irrelevant nodes and edges in the subgraph of the KG and focuses on a sparse set of nodes to make prediction. We examine the nodes that have high signals connection to the target pairs and find literature evidence support. Particularly, the model assigns high weights to two side effects nodes, orthostatic hypotension and aplastic anaemia. Orthostatic hypotension refers to a sudden drop in blood pressure when standing up, and aplastic anemia is a condition in which the body stops producing enough new blood cells. Notably, orthostatic hypotension is closely related to multiple system atrophy, a central nervous system problem~\citep{jones2015orthostatic}. As both drugs incur risk in the side effects, the complication on the central nervous systems could be aggravated when these drugs are taken together, supporting our model prediction. This case study illustrates how to use \mname for potential DDI prediction. }

\subsection{Parameter studies}

We study the effect of  key parameters. When evaluating one parameter, we fix other parameters to their default values.

\begin{figure}[t]
  \centering
  \subfigure[Result of DrugBank]{
    \includegraphics[width=0.23\textwidth]{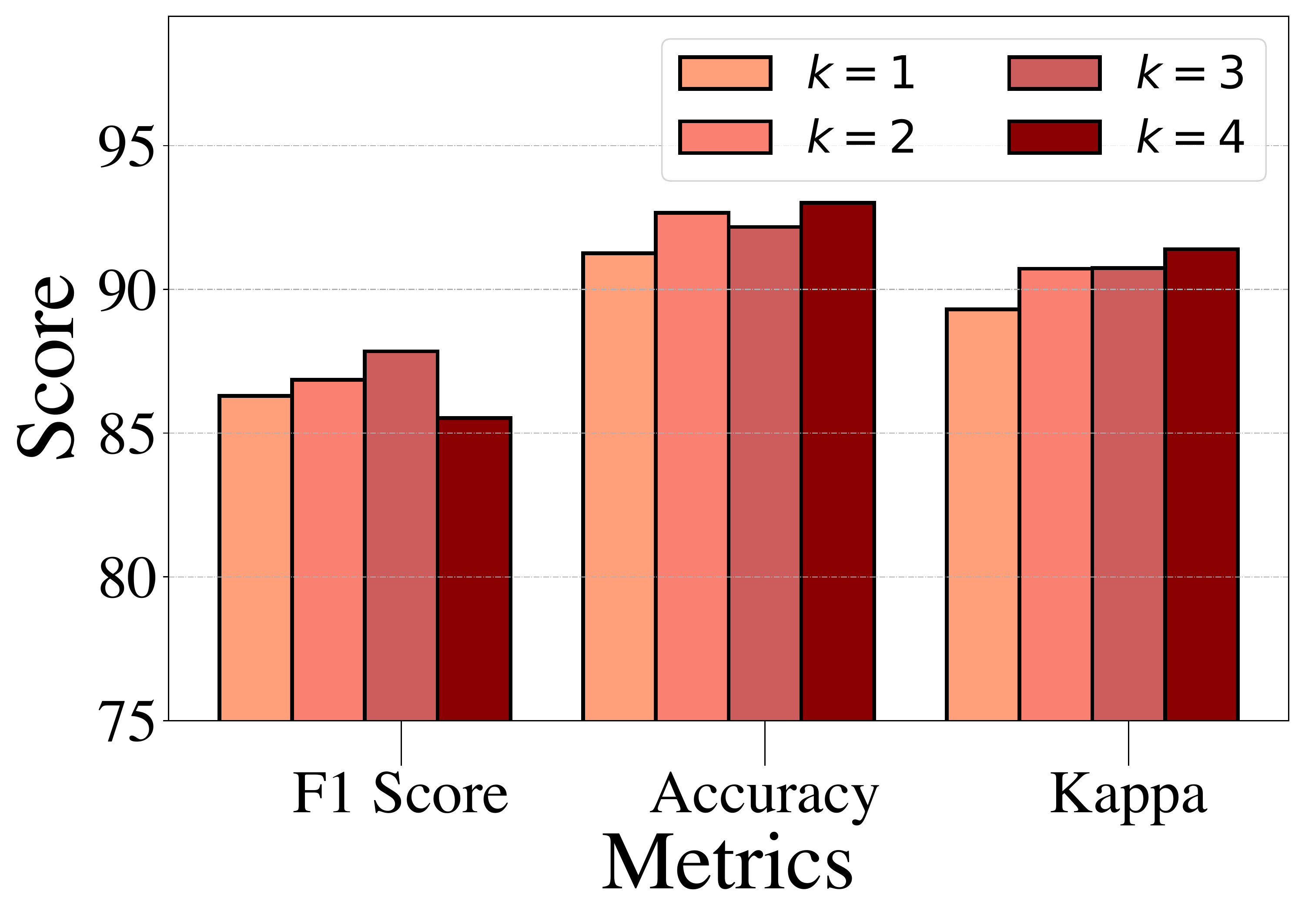}
    \label{fig:lambda}
  }\hspace{-2mm}
  \subfigure[Result of TWOSIDES]{
    \includegraphics[width=0.23\textwidth]{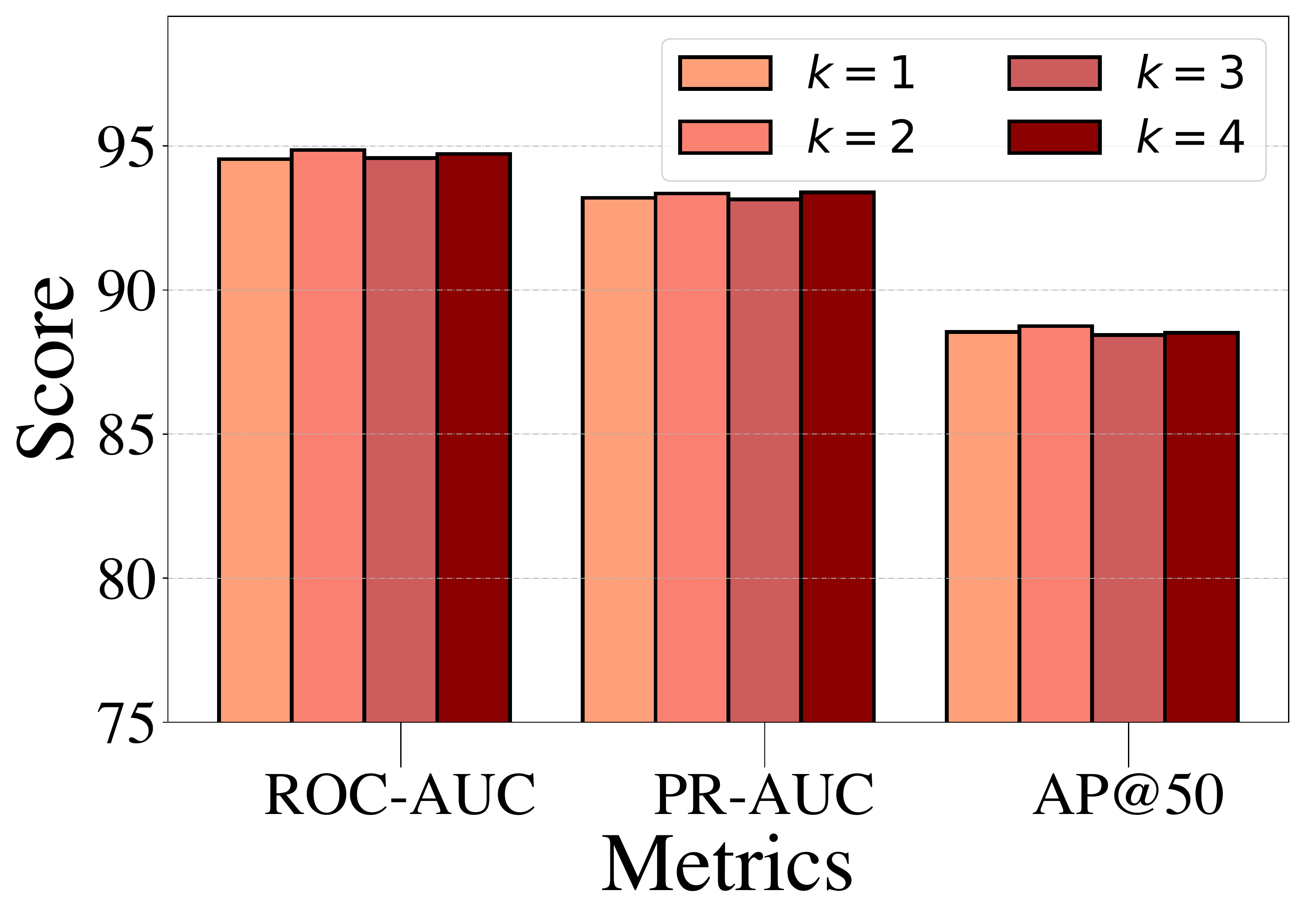}
    \label{fig:mu}
  }
  \vspace{-1mm}
  \caption{Model performance given different parameter $k$.}\label{fig:k}
\end{figure}

\noindent $\bullet$ \textbf{Effect of the hop of the subgraph $k$}: Figure~\ref{fig:k} shows the result of \mname with varying $k$. From the result, we find that for DrugBank dataset, the performance first increase when $k$ is small. However, when $k$ increases from 3 to 4, we observe slight performance drops on F1 score. These indicate that the larger subgraph can bring more useful information while when $k$ is too large, it may also bring some noise and hurt the performance. For TWOSIDES dataset, we find the result is more stable with different $k$, indicating even 1-hop subgraph provides adequate information for DDI task.  

\begin{table}[t]
    \centering
    \caption{The running time for one epoch of \mname for two datasets. Note that \textbf{w/o Subgraph} is a variant that directly aggregates the information for all neighbors on KG.}
    \label{tab:k}
\begin{tabular}{@{}c|c|c|c|c|c@{}}
\toprule
 & \multicolumn{4}{c|}{\textbf{Size of Hop $k$}} &                                         \\ \cmidrule(lr){2-5}
\multirow{-2}{*}{\textbf{Dataset}} & {1}       & 2      & 3      & 4      & \multirow{-2}{*}{\textbf{w/o Subgraph}} \\ \midrule
DrugBank                           & {$132$ s}    & $141$ s   & $178$ s   & $205$ s   & $1279$ s              \\
TWOSIDES                        & {$62$ s}     & $75$ s    & $91$ s    & $102$ s   & $471$ s                                    \\ \bottomrule
\end{tabular}
\end{table}
Moreover, to show how \textbf{subgraph formulation drives efficiency}, we compare the training time between \mname with varying subgraph size and \mname with the entire KG to propagate (See Table~\ref{tab:k}). We find \mname saves 80\% of training time via subgraph anchoring, which demonstrates the efficiency of our approach. 

\begin{figure}[t]
  \centering
  \subfigure[Result of DrugBank]{
    \includegraphics[width=0.23\textwidth]{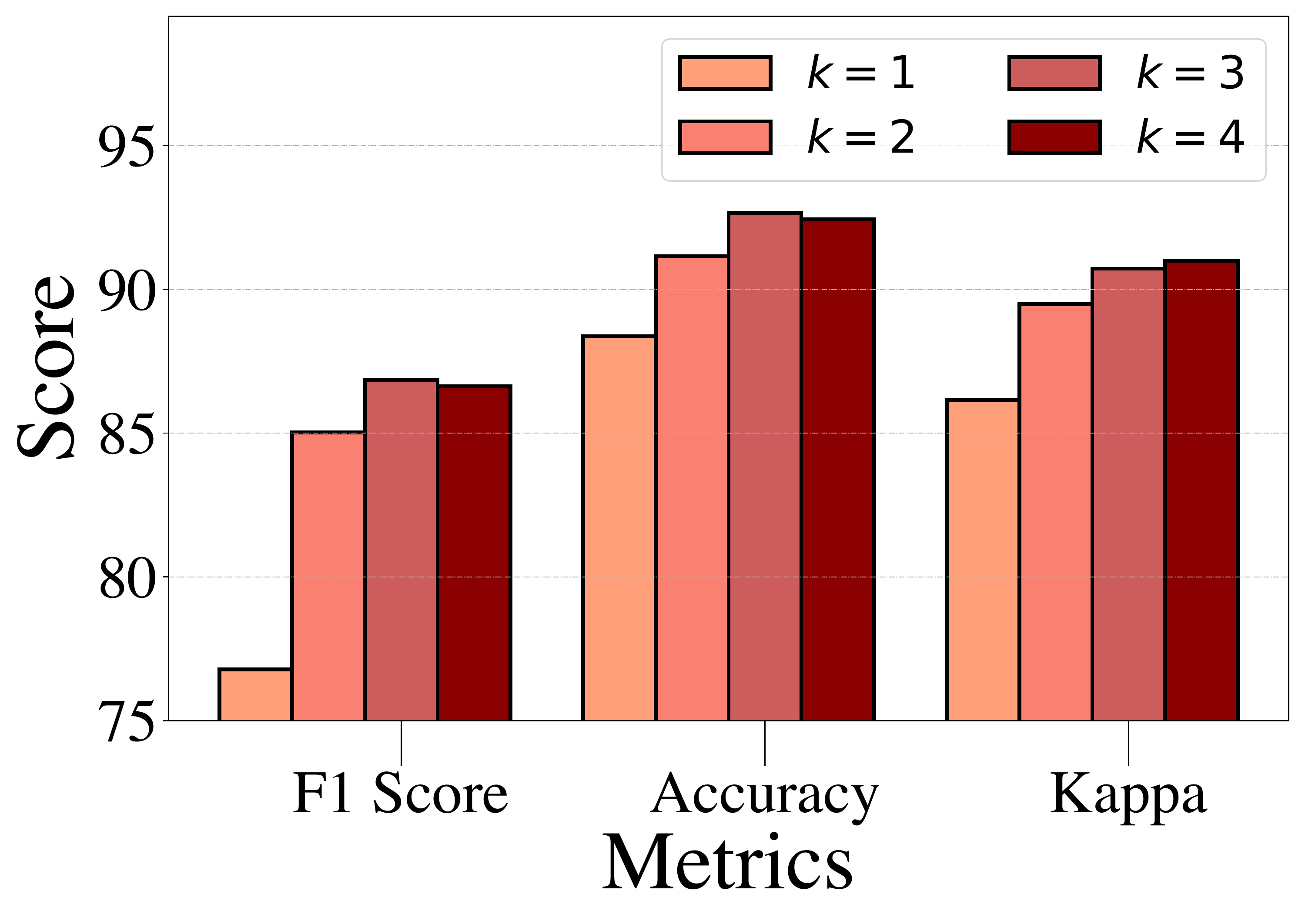}
    \label{fig:lambda1}
  }\hspace{-2mm}
  \subfigure[Result of TWOSIDES]{
    \includegraphics[width=0.23\textwidth]{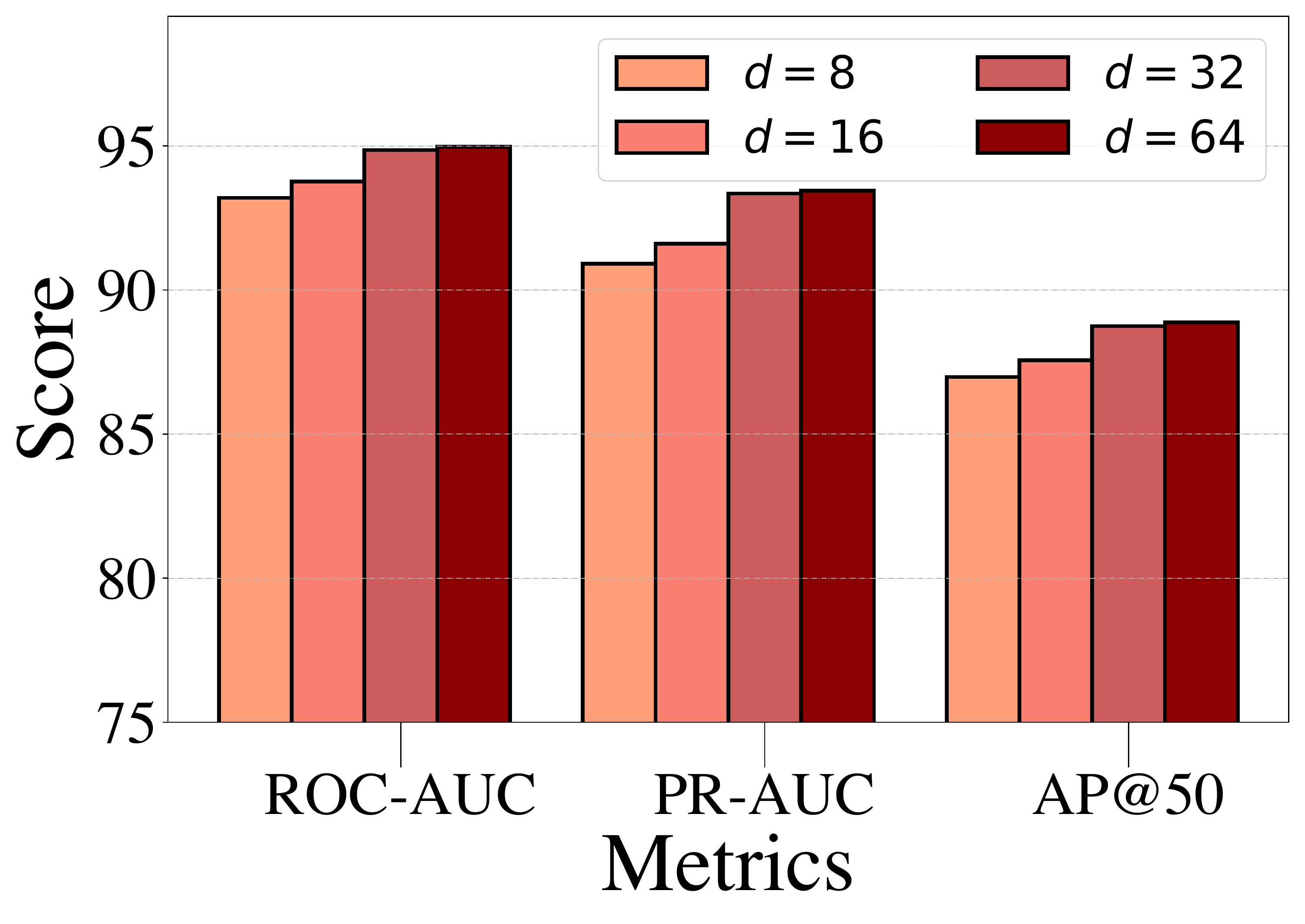}
    \label{fig:mu1}
  }
  \vspace{-1mm}
  \caption{Model performance given different parameter $d$.}\label{fig:d}
   \vspace{-1mm}
\end{figure}

\noindent $\bullet$ \textbf{Effect of the dimension of embeddings $d$}: Figure~\ref{fig:d} exhibits 
the influence of embedding dimension $d$. The result indicates that when $d$ is small, increasing $d$ can boost the performance. But when $d$ becomes large, the gain is marginal. 

\begin{figure}[!ht]
  \centering
  \subfigure[Result of DrugBank]{
    \includegraphics[width=0.23\textwidth]{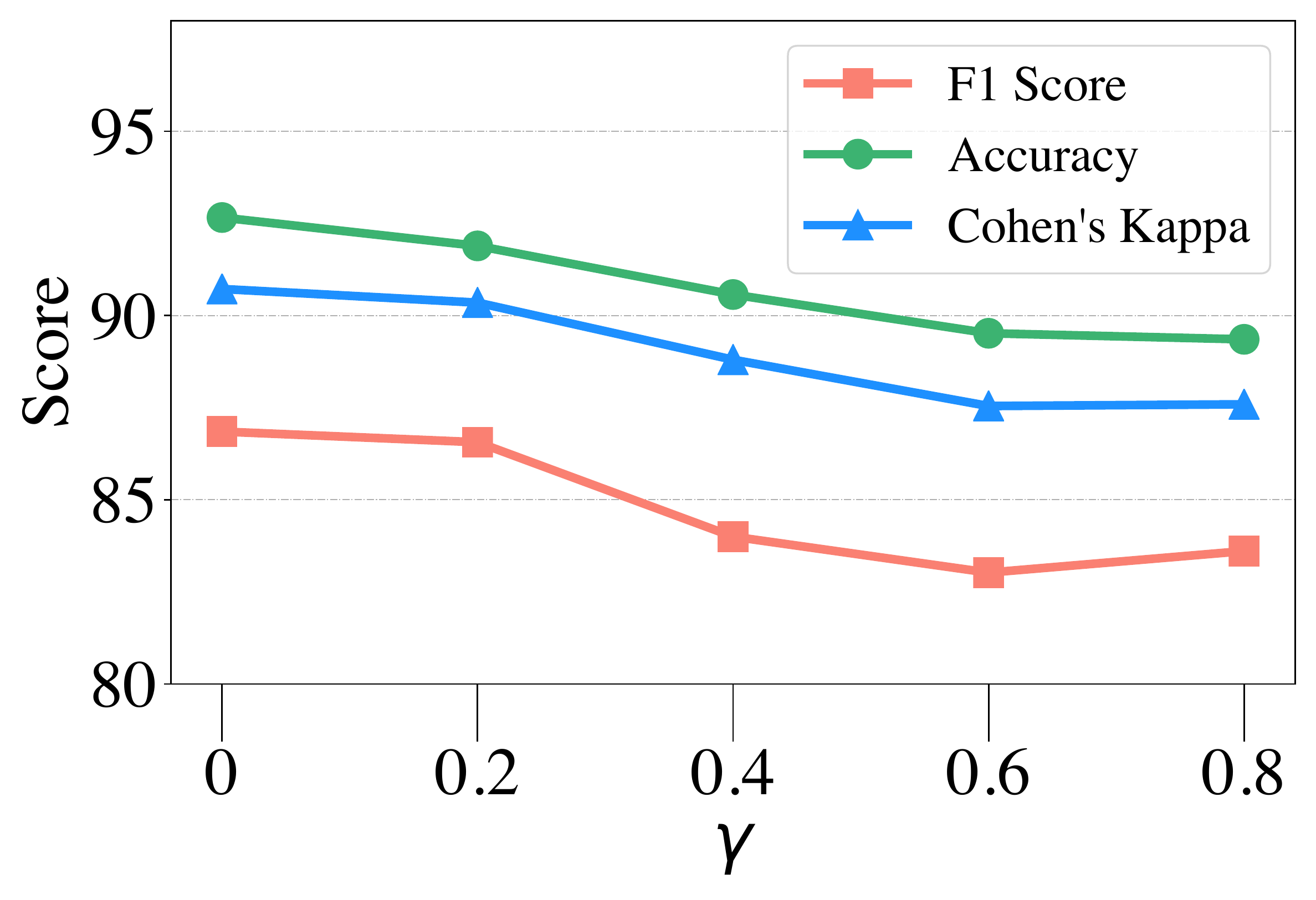}
    \label{fig:lambda2}
  }\hspace{-2mm}
  \subfigure[Result of TWOSIDES]{
    \includegraphics[width=0.23\textwidth]{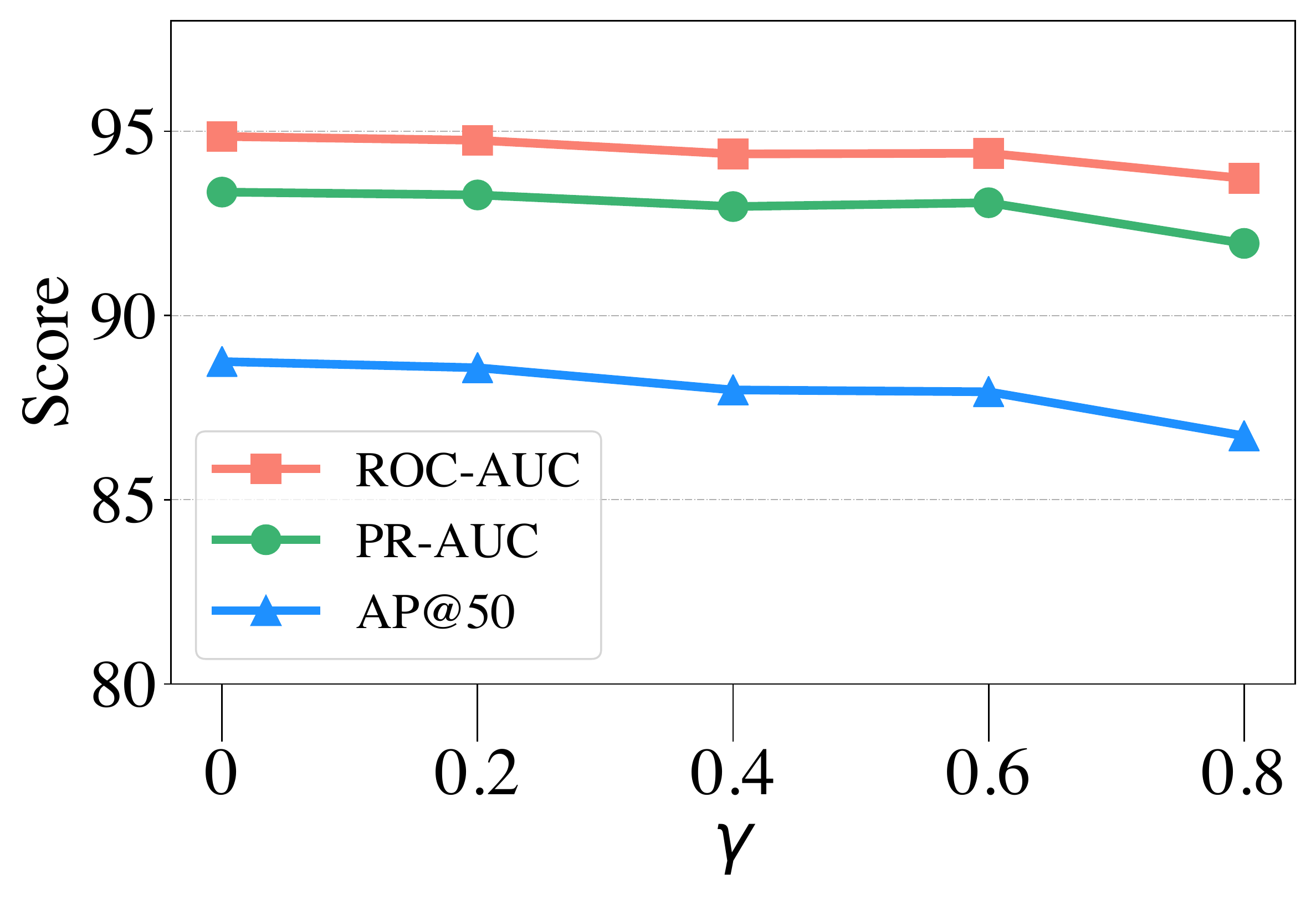}
    \label{fig:mu2}
  }
  \vspace{-1mm}
  \caption{Model performance given different parameter $\gamma$.}\label{fig:gamma}
   \vspace{-1mm}
\end{figure}

\noindent $\bullet$ \textbf{Effect of the threshold for summarization $\gamma$}: Figure~\ref{fig:gamma} shows the result of \mname with different threshold $\gamma$, which demonstrates that when $\gamma$ is small, the performance is rather stable as filtering edges with low score has little effect on the final prediction. This also means \mname is able to achieve similar predictive performance while removing many irrelevant edges. However, when $\gamma$ is large ($\gamma>0.4$ for DrugBank and $\gamma>0.6$ for TWOSIDES), it is clear that the performance drops more. In such cases, the summarized graph is more sparse and we might filter out  potential useful edges. To sum up, there is a trade-off between the explainability and performance.

\section{Conclusion}

In this paper, 
we propose a new method \mname:~{\it knowledge summarization graph neural network} for multi-typed DDI predictions, which is mainly enabled by a local subgraph module that can efficiently anchor on relevant subgraphs from a KG, a self-attention based subgraph summarization scheme that can generate reasoning path within the subgraph, and a multi-channel knowledge and data integration module that utilizes massive external biomedical knowledge for significantly improved multi-typed DDI predictions. Experiments on real-world datasets demonstrated the strong performance of \mname. 


In addition, computational approaches depend heavily on the training data. If the number of training data associated with one specific drug interaction type is low, it is difficult to predict accurately. In contrast to other works, \mname is able to generate good performance in low-resource settings. \mname is also a general framework and can be adapted to predict any other interactions such as drug-disease interaction. The ability to low-resource learning could also mean to excel at finding drugs for rare diseases.


\bibliographystyle{ACM-Reference-Format}
\bibliography{ref}

\clearpage

\appendix
\clearpage

\appendix
\section{Implementation Details}\label{sec:imple}

\subsection{\mname Parameter Setup} We use the following hyperparameter set for \mname after random search on validation set. We use $1,024$-bits Morgan fingerprint for drug featurization. We set the subgraph to be $2$-hops neighbors (i.e. $k=2$). In the subgraph summarization module, we use weight matrix of size $d=32$ for $\mathbf{W}_1$ and $\mathbf{W}_2$. The hidden dimension $\mathbf{h}_v^k$ is set to be $d=32$. The relation matrix $\mathbf{r}$ is set to be 32. The edge pruning threshold is set to be $\gamma = 0$. The input hidden representation of each node is $d=32$. 
The number of basis $B$ in Eq.~\eqref{eq:basis} is set to $8$ as the performance do not change much when set from $4$ to $16$ and suffer from over-fitting with $B>16$. We study the effect of key parameter $d, \gamma$ and $k$ in our experiment part (Section~\ref{experiment}).

\subsection{Training Details}
\textbf{Training Parameters.} For both our method and baselines, the training parameters are set as follows unless specified. \\
We train the model for 50 epochs with batch size 256. Our model is optimized with ADAM optimizer~\citep{kingma2014adam} of learning rate $5\times10^{-3}$ with gradient clipping set to $10$ under L2 norm. We set the  L2 weight decay to $1\times10^{-5}$, the layer of GNN to 2 and set the dropout rate to $0.3$ for each Layer in GNN. \\
\noindent \textbf{Model Implementation and Computing Infrastructure.} 
All methods are implemented in PyTorch\footnote{\url{https://pytorch.org/}} and the graph neural network modules are build on Deep Graph Library~(DGL)\footnote{\url{https://www.dgl.ai/}}.
The System we use is Ubuntu 18.04.3 LTS with Python 3.6, Pytorch 1.2 and DGL 0.4.3. 
Our code is run in a Intel(R) Core(TM) i7-5930K CPU @ 3.50GHz CPU and 
a GeForce GTX TITAN X GPU. 

\subsection{The Range for Tuning Hyper-parameters}
We use grid search to determine hyper-parameters and list the search space of key hyper-parameters as follows.

\begin{table}[h]
 \label{tab:range}
\centering
\caption{The range for tuning hyper-parameters. The bold numbers are the default settings.} 
\begin{tabular}{@{}cc@{}}
\toprule
\textbf{Parameters} & \textbf{Range} \\ \midrule
      Learning Rate     &    $[5\times10^{-4}$, $1\times10^{-3}$, $\mathbf{5}\times\mathbf{10}^\mathbf{-3}, 1\times10^{-2}]$   \\
      Weight Decay     &     $[1\times10^{-6}$, $\mathbf{1}\times\mathbf{10}^\mathbf{-5}, 1\times10^{-4},1\times10^{-3}]$  \\
      Dropout & $[\textbf{0.3}, 0.4, 0.5]$ \\
      Layers of GNN & $[1,\textbf{2},3]$ \\
      $d$ & $[8, 16, \textbf{32}, 64]$ \\ 
      $k$ & $[1, \textbf{2}, 3, 4]$ \\ 
      $B$ & $[4, \textbf{8}, 12, 16, 24, 32]$ \\
      \bottomrule
\end{tabular}
\end{table}

\subsection{Baseline Setup}
For the baselines, the settings are described as follows:
\begin{itemize}
    \item \textbf{MLP}: We implement MLP with Pytorch with the Morgan fingerprint. We use a two-layer MLP and set the hidden dimension to $100$ with dropout $0.3$. 
    \item \textbf{Node2vec}: We follow the officially released implementation from authors\footnote{\url{https://github.com/aditya-grover/node2vec}} and set the embedding dimension to $64$.
    \item \textbf{Decagon}: We use DGL to implement the model. Following~\citep{zitnik2018modeling}, we set the number of GNN layers to $2$ set the hidden dimension to 64 and 32 for two layers with a dropout rate of 0.1 and a minibatch size of 512.
    \item \textbf{GAT}:  We use DGL to implement the model and set the hidden dimension to 64 with 4 attention heads, as we find that improving the number of heads will hurt the performance. We set the activation function to \texttt{LeakyReLU} with $\alpha=0.2$.
    \item \textbf{Others}: 
     We follow the officially released implementa-
tion from the authors listed as follows:
    \begin{itemize}
        \item \textbf{SkipGNN}: {\url{https://github.com/kexinhuang12345/SkipGNN}}.
        \item \textbf{KG-DDI}: the neural model is based on code in  \url{https://github.com/rezacsedu/Drug-Drug-Interaction-Prediction}, and the KG embeddings are trained via OpenKE toolbox \url{https://github.com/thunlp/OpenKE}.
        \item \textbf{GraIL}: \url{https://github.com/kkteru/grail}.
        \item \textbf{KGNN}: \url{https://github.com/xzenglab/KGNN}.
    \end{itemize}
\end{itemize}

\end{document}